%% file: acl_latex.tex
\pdfoutput=1

\documentclass[11pt]{article}

\usepackage[preprint]{acl}

\usepackage{times}
\usepackage{latexsym}
\usepackage{graphicx}
\usepackage{amsmath}
\usepackage{multirow}

\usepackage[T1]{fontenc}

\usepackage[utf8]{inputenc}

\usepackage{microtype}

\usepackage{inconsolata}

\usepackage{graphicx}
\usepackage{amssymb}

\DeclareMathSymbol{\minus}{\mathbin}{AMSa}{"39}

\usepackage[inline]{enumitem}
\usepackage{cleveref}
\usepackage{float}

\title{Extending Context Window of Large Language Models from a Distributional Perspective}

\author{\textbf{Yingsheng Wu$^1$\thanks{~Equal Contribution},} \bf \textbf{Yuxuan Gu$^1$\footnotemark[1],} \bf \textbf{Xiaocheng Feng$^1$,} \bf \textbf{Weihong Zhong$^1$,} \\
\textbf{Dongliang Xu$^2$,} \bf \textbf{Qing Yang$^2$,} \bf \textbf{Hongtao Liu,$^2$} \bf \textbf{Bing Qin$^1$} \\
  $^1$Harbin Institute of Technology, Harbin, China \\
  $^2$Du Xiaoman (Beijing) Science Technology Co., Ltd. \\
  \texttt{\{yswu,yxgu,xcfeng,whzhong,qinb\}}@ir.hit.edu.cn \\
  \texttt{\{xudongliang,yangqing,liuhongtao\}@duxiaoman.com}
  }

\begin{document}
\maketitle
\input{section/1_abstract}
\input{section/2_intro}

\input{section/3_background}

\input{section/4_method}

\input{section/5_experiment}

\input{section/6_analysis}
\input{section/8_relatedworks}
\input{section/7_conclusion}

\section{Limitations}
Our method is limited by the rotary position embedding, which is not currently available for LLMs with other embedding methods. However, this is not a serious problem because (1) the most powerful open source LLMs, such as LLaMA2, utilize the rotary position embedding, and (2) our approach addresses the problem from a theoretical perspective, which can be better generalized to other embedding frameworks in future research than empirical work.

When applying the model to long contextual tasks, the quadratic computational complexity problem of transformers still exists. Fortunately, our method does not introduce more computational overhead in the inference phase. Besides, we are compatible with other computationally efficient Transformer methods.

Our method does not make any structural improvements to the rotation position embedding or interpolation methods, so it still does not fully achieve the optimal situation with the distribution perturbation $\mathcal{D}(P_{L'},P_{L})=0$. This provides inspiration for future exploration.

The accuracy of our estimated rotary angle distribution is affected by the pre-training sequence length $L$, since the rotary angles are regarded as sampled $L$ times from the real rotary angle distribution. Currently, our method can achieve satisfying improvement for models with $L=4\text{k}$, and will perform better when applied for models with longer pre-training length.

Due to the constraints of computing resources, our experiments are limited to LLaMA2-7B and LLaMA2-13B, and the long contextual ability is also constrained by the model size. In the future, we hope to apply our method to extend the context window of even larger models to achieve stronger long contextual abilities.

\section{Ethics Statement}
We are totally aware that text generation technology has a potential to be used maliciously to generate fake, toxic, or offensive content. 
We are aware that if LLMs generate harmful or toxic information, our approach cannot explicitly prevent it.
However, since the models and datasets used in our study are publicly available and examined, we are confident that our approach will not introduce toxic content during the length extension phase.
\section{Acknowledgments}
Xiaocheng Feng is the corresponding author of this work. We thank the anonymous reviewers for their insightful comments. This work was supported by the National Natural Science Foundation of China (NSFC) (U22B2059, grant 62276078), the Key R\&D Program of Heilongjiang via grant 2022ZX01A32, the International Cooperation Project of PCL, PCL2022D01 and the Fundamental Research Funds for the Central Universities (Grant No.HIT.OCEF.2023018).

\bibliography{ref}
\clearpage
\appendix

\section{Rotation Angle Distribution Details}
\label{sec:appendixA}
\subsection{Rotation Angle Distribution}
\label{sec:appendixA.1}
\begin{figure}[h]
    \centering
    \includegraphics[width=1\linewidth,height=!]{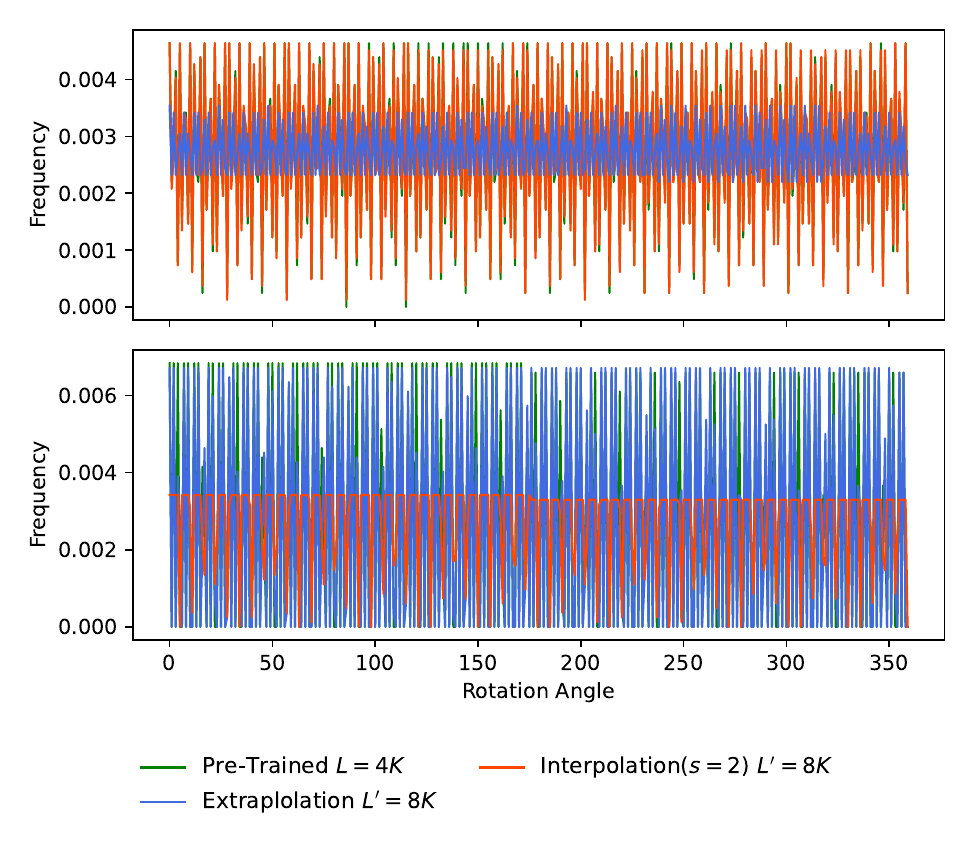}
    \caption{Complete rotary angle distributions of 6th and 22nd dimensions when the number of intervals is set to 360.}
    \label{distribution_complete}
\end{figure}

Figure ~\ref{distribution_complete} illustrates the complete rotary angle distributions of the 6th and 22nd dimensions when the number of intervals is set to 360.
\subsection{Disturbance of Different Method}
\label{sec:appendixA.2}
\begin{figure}[h]
    \centering
    \includegraphics[width=1\linewidth,height=!]{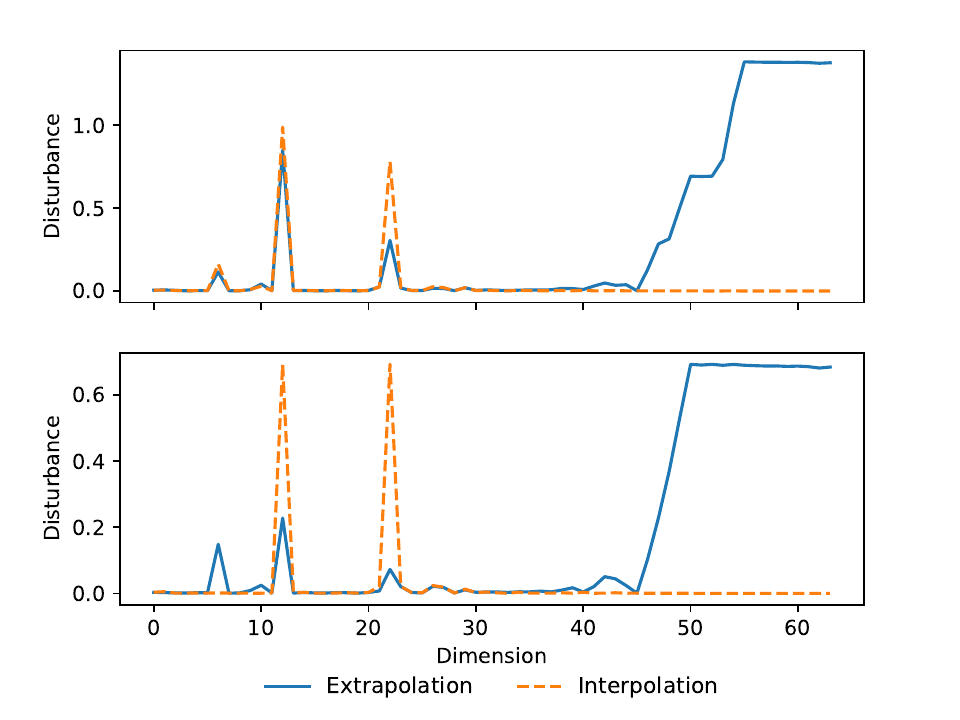}
    \caption{Illustration of the impact of interpolation and extrapolation on each dimensional distribution. \textbf{Upper}: Disturbance when the context window is extended to 8k. \textbf{Lower}: Disturbance when the context window is extended to 16k.}
    \label{fig:ext_int}
\end{figure}

~\Cref{fig:ext_int} illustrates the disturbance to each dimensional distribution caused by interpolation and extrapolation when the context window of the model is extended to 8k and 16k. Interpolation and extrapolation exhibit advantages in different dimensions, respectively.

\begin{figure}[h]
    \centering
    \includegraphics[width=1\linewidth,height=!]{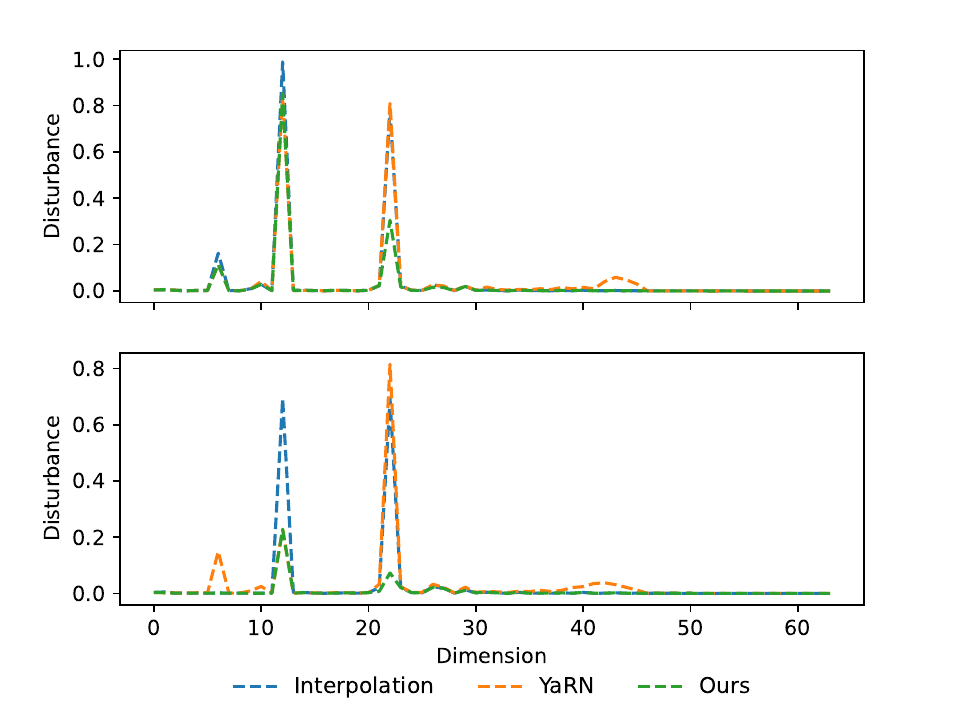}
    \caption{Illustration of the impact of PI, YaRN and our method on each dimensional distribution. \textbf{Upper}: Disturbance when the context window is extended to 8k. \textbf{Lower}: Disturbance when the context window is extended to 16k.}
    \label{fig:yarn_ours}
\end{figure}

~\Cref{fig:yarn_ours} illustrates the disturbance to each dimensional distribution caused by PI\citep{PI} ,YaRN\citep{yarn} and our method when the context window of the model is extended to 8k and 16k. Our method achieves the lowest disturbance to the distribution. 
\section{Experimental Details}
\label{sec:appendixB}
\subsection{Experimental Setup}
\label{sec:appendixB.1}
We use 8 A100 GPUs and adopt ZeRO3 \citep{zero} strategies during the training stage, and use AdamW \citep{adamw} optimizer with $\beta_1$ = 0.9 and $\beta_2$ = 0.999. We set the learning rate to 2 × 10$^{-5}$ without warmup and weight decay. 
When extending the context window to 8k, we spent approximately 6 hours training LLaMA-7B and approximately 10 hours training LLaMA2-13B. When extending the context window to 16k, we spent approximately 7 hours training LLaMA-7B and approximately 11 hours training LLaMA2-13B.
Both training and testing are accelerated by FlashAttention-2 \citep{flashattention2}.

\subsection{Additional Experimental Results}
\label{sec:appendixB.2}
\input{table/ruler}
\subsubsection{RULER Benchmark}
The RULER \citep{RULER} benchmark is employed to evaluate the long-context retrieval capabilities of models, with the performance of different methods on this benchmark presented in Table ~\ref{tab:ruler_main}. Although the retrieval performance on short texts has decreased, all methods have enhanced the model's ability to retrieve information from long documents, with our approach achieving the highest retrieval accuracy. The original LLaMA2 model, due to its limited capacity for handling long documents, fails to produce accurate answers when the context length exceeds 4k tokens. The inferior performance of CLEX may be attributed to the introduction of new parameters for predicting the scaling factor, which requires more training data to fit, thereby leading to sub-optimal performance in scenarios with limited data.

\subsubsection{Time complexity}
Considering the balance between efficiency and performance, we also provide the time consumption of different methods, as shown in Table ~\ref{tab:time}. To facilitate comparison, we normalized the time consumption. In comparison to a fixed scaling factor, CLEX introduces additional parameters to predict the scaling factor, which necessitates the recalculation of positional encoding, thereby increasing the training and inference times.
\input{table/time}

\subsubsection{Perplexity}
Perplexity is commonly employed to evaluate a model's language modeling capabilities, and we tested the perplexity of different methods under non-training conditions, with the results presented in Table ~\ref{tab:ppl}. However, perplexity often fails to reflect a model's actual performance on downstream tasks, as a model may exhibit a relatively low perplexity in non-training scenarios yet perform poorly in real-world applications. In contrast to the decrease in perplexity, we are more concerned with the model's performance on actual tasks.

\input{table/ppl}

\subsection{Passkey Prompt}
\label{sec:appendixB.3}
We follow experimental setup of \citet{passkey, PI}. We separately employed our method with scaling factors of $s$=2 and $s$=4 to extend the context windows of LLaMA2 7B and 13B to 8k and 16k, respectively. Figure ~\ref{fig:passcode-prompt} shows the prompt template. 
\begin{figure}[H]
    \small
	\texttt{\noindent
		There is an important info hidden inside a lot of irrelevant text. Find it and memorize them. I will quiz you about the important information there.\\
		The grass is green. The sky is blue. The sun is yellow. Here we go. There and back again. \underline{(repeat n times)} \\
		The pass key is \textbf{12345}. Remember it. \textbf{12345} is the pass key.\\
		The grass is green. The sky is blue. The sun is yellow. Here we go. There and back again. \underline{(repeat m times)} \\
		What is the pass key? The pass key is\\
	}
	\caption{\small Prompt format for passkey retrieval. Here the passkey 12345 is replaced with a random 5-digit numbers during test and the prompt length varies with the value of n and m.}
	\label{fig:passcode-prompt}
\end{figure}

\end{document}

%% file: section/1_abstract.tex
\begin{abstract}
Scaling the rotary position embedding (RoPE) has become a common method for extending the context window of RoPE-based large language models (LLMs). However, existing scaling methods often rely on empirical approaches and lack a profound understanding of the internal distribution within RoPE, resulting in suboptimal performance in extending the context window length. 
In this paper, we propose to optimize the context window extending task from the view of rotary angle distribution. 
Specifically, we first estimate the distribution of the rotary angles within the model and analyze the extent to which length extension perturbs this distribution.
Then, we present a novel extension strategy that minimizes the disturbance between rotary angle distributions to maintain consistency with the pre-training phase, enhancing the model's capability to generalize to longer sequences. Experimental results compared to the strong baseline methods demonstrate that our approach reduces by up to 72\% of the distributional disturbance when extending LLaMA2's context window to 8k, and reduces by up to 32\% when extending to 16k. On the LongBench-E benchmark, our method achieves an average improvement of up to 4.33\% over existing state-of-the-art methods. Furthermore, our method maintains the model's performance on the Hugging Face Open LLM benchmark after context window extension, with only an average performance fluctuation ranging from -0.12 to +0.22. Our code is available at \url{https://github.com/1180301012/DPRoPE}.

\end{abstract}

%% file: section/2_intro.tex
\begin{figure}[!t]
    \centering
    \includegraphics[width=1\linewidth,height=!]{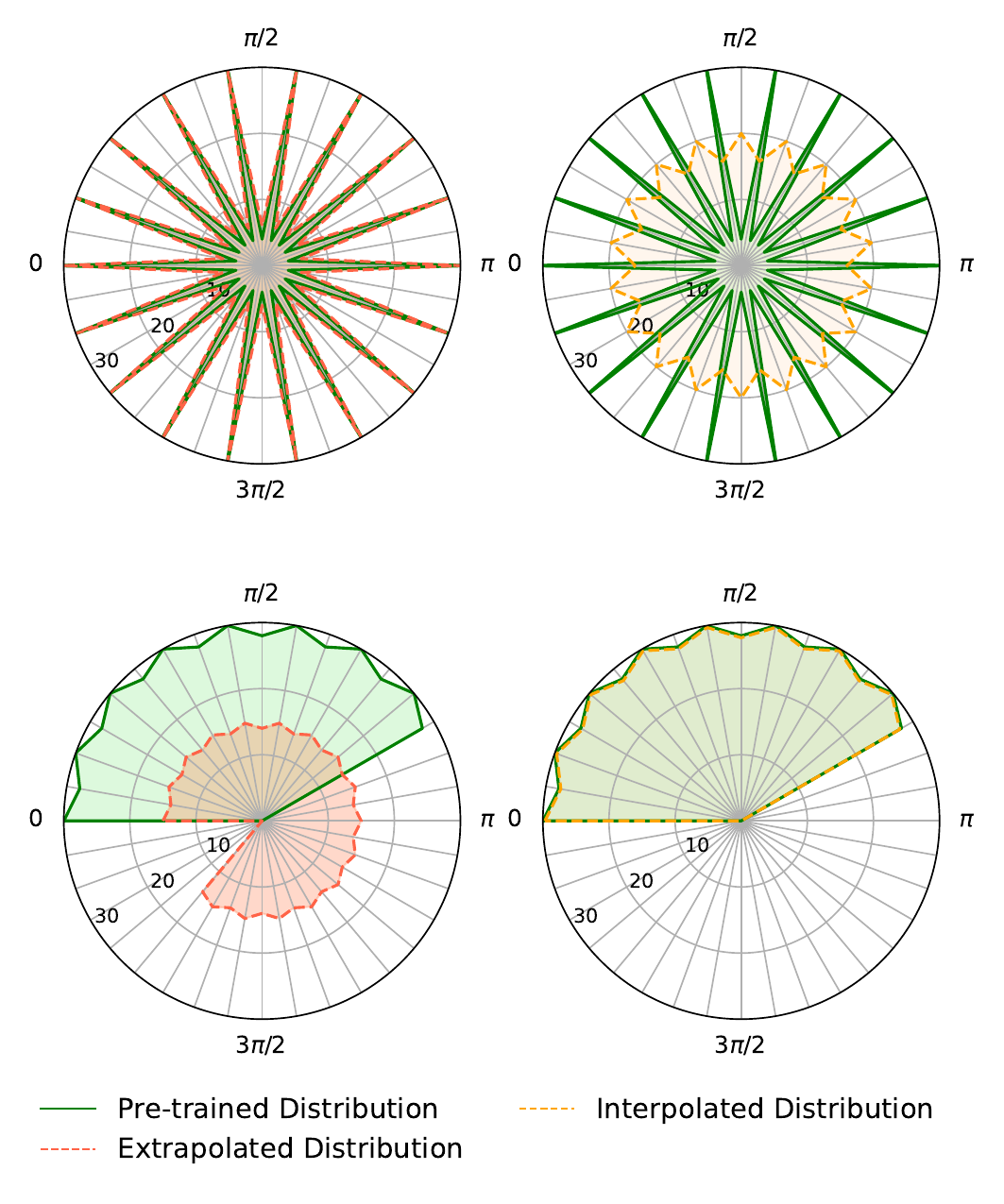}
    \caption{Rotary angle distributions of extrapolation and interpolation methods in two different dimensions, compared with the pre-trained angle distribution. (a) In one dimension, the extrapolated rotary angle distribution fits more closely with the pre-trained distribution. (b) In another dimension, the interpolated distribution fits better with the pre-trained distribution.}
    \label{fig:angle_distribution}
\end{figure}
%
\section{Introduction}
Given the remarkable capabilities of transformer-based large language models (LLMs) in addressing a wide range of natural language processing tasks \citep{gpt4-report, llama, llama2, mixtral}, modeling arbitrarily long textual sequences remains a significant challenge. On the one hand, LLMs trained on short sequences often encounter out-of-distribution (OOD) issues when applied to the longer ones \citep{scalinglaw}. On the other hand, training an LLM with extremely long context windows (i.e., the maximal sequence length) from scratch is expensive and inefficient. Currently, the most popular approach is pre-training a large language model, such as LLaMA, Qwen2 \citep{llama, llama2, qwen2}, with a limited context window and the rotary position embedding (RoPE, \citet{rope}). During the inference stage, the context window is dynamically extended via fine-tuning or tuning-free position interpolation strategies \citep{PI, yarn, scalinglaw} on the rotary position embedding.


However, these position interpolation strategies primarily rely on intuition and are developed from an empirical perspective, resulting in a lack of interpretability \citep{lesurvey} and sub-optimal performance for context extension. 
For example, PI \citep{PI} equally stretches all dimensions of the RoPE with the context extension ratio. YaRN \citep{yarn} observes that heuristically utilizing different strategies for different dimensions yields better performance. However, the reasons behind this phenomenon have not been thoroughly investigated, resulting in it likely not achieving the best results. Moreover, the optimal hyperparameters determined experimentally in YaRN potentially hinder its generalization to new model settings.

To bridge the gap between experiments and theoretical analysis, we tackle context window extension from the view of rotary angle distribution. Hence, we propose a method for length extension strategy selection, which has the potential to be theoretically optimal by minimizing the perturbation to the rotary angle distributions of the pre-trained language model. 
Specifically, we first compare the pre-training rotary angle distribution with the distributions introduced by interpolation and extrapolation. As illustrated in \Cref{fig:angle_distribution}(a), interpolation can introduce too many OOD angles that have a frequency of 0 in the pre-training distribution, indicating a significant disturbance to the original distribution and posing a challenge for the model to adapt to the new distribution.  While direct extrapolation may have a negligible impact on the distribution. Contrarily in another dimension demonstrated in \Cref{fig:angle_distribution}(b), direct extrapolation introduces numerous OOD angles in this situation, causing a severe distribution disturbance, whereas interpolation performs better.  

From such distributional view, we find that the consistency between the pre-training rotary angle distribution and the extension distribution varies across different dimensions. Thus, we propose to employ different extension strategies in different dimensions according to the rotary angle distribution. 
We first approximate the distributions of rotary angles by calculating the frequency of angles in minimal discrete intervals. 
Then, we estimate the disturbance introduced by different extension strategies by computing the distance between the interpolated or extrapolated distribution and the original one. 
Finally, we determine the most appropriate extension strategy for each rotary angle dimension independently.



Experiments across LLMs of different sizes and various long-context tasks demonstrate the effectiveness of our distributional approach. We outperform the strong extension baselines PI \citep{PI} and YaRN \citep{yarn} on LongBench-E \citep{longbench}, achieving a new state-of-the-art. Besides, our method achieves 100\% accuracy on passkey retrieval \citep{passkey} and matches the performance of original LLMs on short-text tasks in the HuggingFace Open LLM Leaderboard \citep{huggingface}. 
In summary, our contributions are as follows:
\begin{itemize}
    \item We are the first, to the best of our knowledge, to analyze the context window extension from a distributional perspective, where rotary angle distributions are observed to be crucial.
    \item We propose a novel method to minimize the perturbation to the distribution when applying position interpolation for context extension.
    \item Experimental results demonstrate that we can surpass existing long-text extension methods on both long-text and short-text benchmarks.
\end{itemize}

%% file: section/3_background.tex
\section{Preliminaries}
\subsection{Rotary Position Embedding (RoPE)}
\begin{figure}[t]
    \centering
    \includegraphics[width=1\linewidth,height=!]{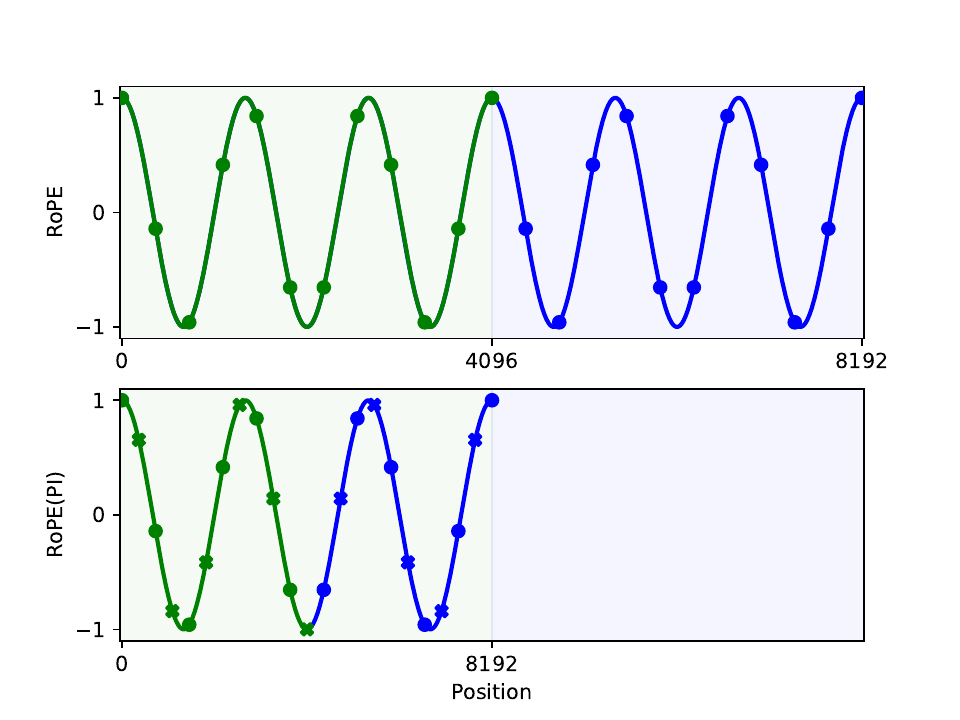}
    \caption{An example of context window extension, where green and blue points denote pre-trained and OOD position indices. \textbf{Upper}: Extrapolation directly models position indices with RoPE. \textbf{Lower}: Interpolation mitigates the OOD problem of position indices while introducing unseen rotary angles (cross points).}
    \label{fig:PI}
\end{figure}

Rotary position embedding \citep{rope} is a position embedding method widely used in recent LLMs, which have weak extrapolation properties for long text modeling and context window extension.
As demonstrated in the upper part of \Cref{fig:PI}, OOD position indices can be directly extrapolated when corresponding rotary angles are periodic.
Given a $d$-dimensional attention head, the $m$th token's rotary matrix $\mathcal{R}_m^d$ is defined:
\begin{equation}
    \label{equ:theta}
    \mathcal{R}_m^d = \begin{bmatrix}
        ..& .. & 0& 0 & 0& 0\\
        ..& ..& 0& 0 & 0& 0\\
        0& 0 & \cos(m\theta_i)& \minus\sin(m\theta_i) & 0&0\\
        0& 0 & \sin(m\theta_i)& \cos(m\theta_i) & 0& 0\\
        0& 0 & 0 & 0 & .. & .. \\
        0& 0 & 0 & 0 & .. & .. \\
    \end{bmatrix}
\end{equation}
where $i\in [0,d/2-1]$ and $\theta_i = 10000^{\minus\frac{2i}{d}}$, where the hyperparameter $10000$ is the default base of RoPE \citep{rope}. 
Suppose the input of a single attention head is $x_1,\cdots, x_l \in \mathbb{R}^d$, where $l$ is the sequence length and $d$ is the dimension of an attention head. With trainable parameters $\mathbf{W}_q$ and $\mathbf{W}_k$, the the attention logit $\mathbf{q}^\top_m\mathbf{k}_n$ with RoPE can be calculate as follows:
\begin{equation}
    \begin{aligned}
        \mathbf{q}^\top_m\mathbf{k}_n &= (\mathcal{R}_m^d\mathbf{W}_qx_m)^\top(\mathcal{R}_n^d\textbf{W}_kx_n) \\
        &= x_m^\top\textbf{W}_q\mathcal{R}_{n-m}^d\mathbf{W}_kx_n
    \end{aligned}
\end{equation}
where $\mathcal{R}_{n-m}^d=(\mathcal{R}_m^d)^T\mathcal{R}_n^d$ \citep{rope}.
\subsection{Position Interpolation (PI)}
As shown in the lower part of \Cref{fig:PI}, PI \citep{PI} suggests linear interpolation to all dimensions to keep position indices within the pre-trained range. When extending the context window from $L$ to $L'$, with the scaling factor $s=L'/L$, the new $\hat{\theta}_i$ is scaled correspondingly as $\hat{\theta}_i=\theta_i/s$. Although alleviating OOD position indices, this approach is likely to disturb the original periodicity and add unseen rotary angles.
\subsection{YaRN}
For each dimension pair $(2i,2i+1)$ in RoPE, \citet{yarn} define its wavelength as follows:
\begin{equation}
    \lambda_{2i} = \lambda_{2i+1} = 2\pi/\theta_i = 2\pi\cdot10000^{\frac{2i}{d}}.
\end{equation}
YaRN \citep{yarn} argues that high-frequency dimensions should employ less scaling, significantly improving the performance of positional interpolation.
They introduce the ratio $r_i$ between the original context size $L$ and the wavelength $\lambda_i$, which is $r_i = L/\lambda_i$, and apply different scaling strategies to each dimension according to $r_i$.
Given two threshold hyperparameters $\alpha, \beta$, YaRN modifies the RoPE as follows:
\begin{equation}
    \hat{\theta}_i = \begin{cases}
        \theta_i/s, & \text{if} \ \ r_i < \alpha \\ 
        \theta_i, & \text{if} \ \ r_i > \beta\\ 
        (1-\gamma_i)\theta_i/s + \gamma_i \theta_i, & \text{otherwise} \\
    \end{cases},
    \label{equ:yarntheta}
\end{equation}
where $s$ is the scaling factor and $\gamma_i = (r_i-\alpha)/(\beta-\alpha)$.
As shown in \cref{equ:yarntheta}, extrapolation is used for high-frequency dimensions ($r_i>\beta$), while interpolation is used for low-frequency dimensions ($r_i<\alpha$). Others are deployed with NTK-aware \citep{fixedNTK, dynamicNTK} methods.
\citet{yarn} empirically suggest $\alpha=1$ and $\beta=32$ for LLaMAs.

%% file: section/4_method.tex
\section{Method}
In this section, we first introduce how to estimate the rotary angle distribution. Then, we propose a novel approach that extends the context window of LLMs by minimizing the disturbance of the rotary angle distribution.

\subsection{Rotary Angle Distribution}
LLMs generate language sequences by sampling from the learned distribution $p(x)=\prod_m p(x_m|x_{<m})$, where the position order is implicitly controlled by position embedding.
This means that changes in the distribution of position embedding will have an impact on the language distribution.
Thus, we need to model this distribution and maintain its consistency when extending the context window.

As illustrated in \cref{equ:theta}, rotary angles $\Theta_m^i=(m\theta_i \bmod 2\pi)$ of a specific dimension $i$ are finite discrete numbers during the pre-training stage, since $0\leq m< L, m \in \mathbb{N}$. 
Considering them as sampled from the rotary angle distribution, we can statistically estimate this distribution.
We divide the rotary range $[0,2\pi)$ uniformly into $b$ intervals, where the $k$th interval in $i$th dimension is defined:
\begin{equation}
    \text{Interval}^i_k = \left[\frac{2k\pi}{b},\frac{2(k+1)\pi}{b}\right),
     \label{equ:bucket}
\end{equation}
where $k=0,\dots,b-1$, we set the default value of $b$ to 360. 
The frequency of rotary angles $F^i_k(L)$ in each interval is calculated as:
\begin{equation}
    F_k^i(L) = \left\lvert\left\{\Theta_m^i \in \text{Interval}^i_k,\ \forall m \in [0,L)\right\}\right\rvert \big/ L.
\end{equation}
Therefore, the discrete probability density function of rotary angle distribution at the $i$th dimension is:
\begin{equation}
    P^i_L(\Theta \in \text{Interval}^i_k) = F_k^i(L),
\end{equation}
where there is $\sum\limits_{k=0}^{b-1} P^i_L(\Theta \in \text{Interval}^i_k) = 1$.

\begin{figure}[t]
    \centering
    \includegraphics[width=1\linewidth,height=!]{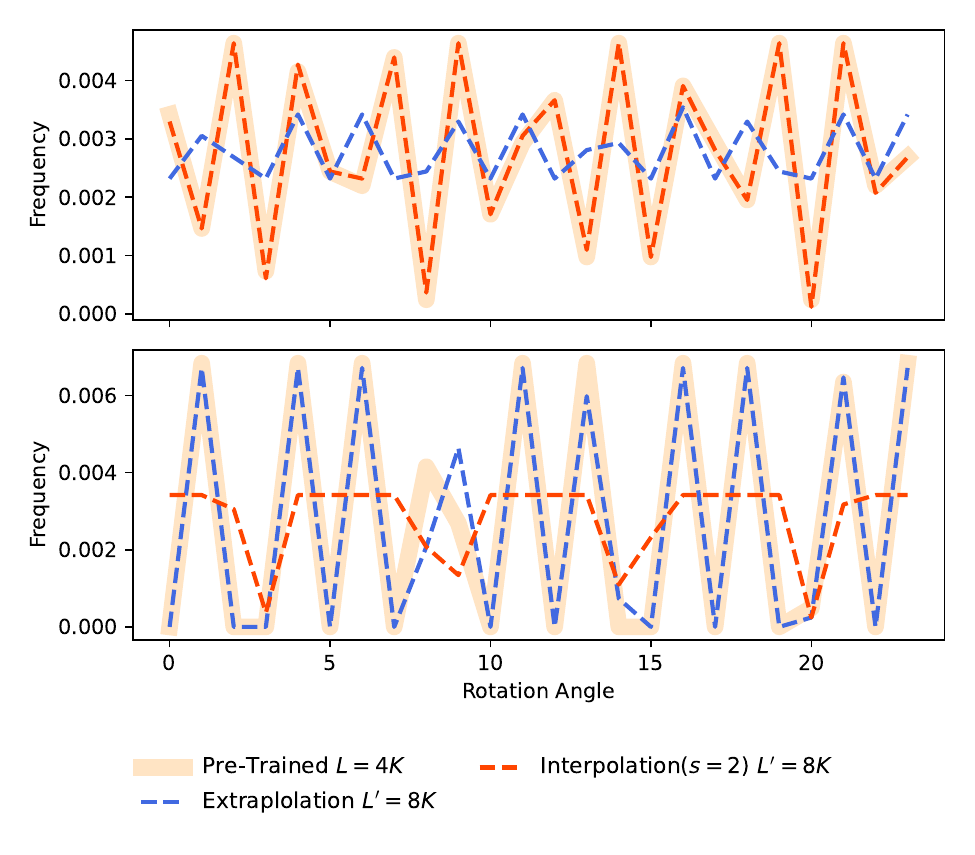}
    \caption{The learned rotary angle distributions of LLaMA2. We demonstrate the $6$th and $22$nd dimensions during pre-training within the 4k length, and the corresponding rotary angle distributions when extended to 8k via interpolation and extrapolation, respectively. We set the number of intervals to $b=360$ and we only display the first 24 intervals for clarity. The distributions of full intervals are provided in \cref{sec:appendixA.1}.}
    \label{fig:frequency}
\end{figure}

Take LLaMA2-7B as an example, where $L=4k$ and $d=128$, we analyze the rotary angle distribution of pre-trained parameters.
We demonstrate the distributions in \Cref{fig:frequency}, which vary significantly as the dimension changes.
When extending the context window to $L'$, such as $L'=8k$, we consider two scenarios for each dimension: interpolation with the scaling factor $s=2$ and direct extrapolation.
Consistency of the distributions derived by these two extension approaches with the original distribution also changes with different dimensions. As shown in \Cref{fig:frequency}, the rotary angle distribution of the interpolation enables better maintenance of consistency with the pre-trained distribution on the $6$th dimension. When it comes to the $22$nd dimension, the situation is completely the opposite. 
Furthermore, we observe that interpolation introduces too many OOD angles that are assigned the frequency of $0$ by the pre-trained distribution, challenging model's generalization capability.

It's worth noting that our observation is inline with the empirical strategies in YaRN \citep{yarn}, where different dimensions have completely different situations.
Besides, distributional consistency is essential for mitigating the OOD issue, which enables LLMs to generalize to longer context window and improves its performance on long-text tasks.
Therefore, we will choose the context window extension methods with the least perturbation according to the rotary angle distribution on different dimensions.

\subsection{Minimizing Distribution Disturbance}
In this part, we derive the disturbance between rotary angle distributions and minimizing the disturbance to maintain the their consistency.
Given a LLM pre-trained on the sequence length of $L$ with the rotary position embedding, the set of rotary angle distributions for all dimensions is denoted as $P_L = \left\{P^0_L(\Theta),\dots, P_L^{d/2-1}(\Theta)\right\}$.
Extending the context window to $L'$, the new rotary angle distribution set is $P_{L'}$.
We define the disturbance $\mathcal{D}(L',L)$ between these two distributions $P_{L'}$ and $P_L$ as:
\begin{equation}
    \begin{aligned}
    \mathcal{D}^i(P_{L'},P_{L}) &= \sum_{k=0}^{b-1} F_k^i(L') \log \frac{F_k^i(L')+\epsilon}{F_k^i(L)+\epsilon} \\
    \mathcal{D}(P_{L'},P_{L}) &= 2\times\sum\limits_{i=0}^{d/2-1} \mathcal{D}^i(P_{L'},P_{L})\Big/d,
    \end{aligned}
    \label{equ:distance}
\end{equation}
where $\epsilon$ is an extremely small number to prevent dividing $0$ and $D^i(P_{L'},P_{L})$ is the KL divergence. For OOD rotary angles introduced by interpolation or extrapolation, $D^i(P_{L'},P_{L})$ yields a high disturbance score due to the large value of $F_k^i(L')$. The score is low when $F_k^i(L')\ll F_k^i(L)$, since the incomplete sampling from the pre-trained rotary angle distribution does not have a serious impact during the inference stage.

Now we can quantitatively compare the situation in \Cref{fig:frequency} and we can further control the extension strategy in a fine-grained manner with the disturbance score, where the primary objective is to minimize the disturbance, $\min \mathcal{D}(P_{L'},P_{L})$.
\input{table/longbench_main}
In detail, we combine the two strategies: one is based on PI, where we use $s=L'/L$ to interpolate and obtain the corresponding rotary angle distributions $P_{L'}^\mathcal{I}$, and the other involves directly extrapolating to $L'$ with distributions $P_{L'}^\mathcal{E}$.
We minimize the disturbance score for each dimension independently, since $\min \mathcal{D}(P_{L'},P_{L}) \propto \sum\limits_{i=0}^{d/2-1} \min \mathcal{D}^i(P_{L'},P_{L})$, via selecting interpolation or extrapolation based on the score.
Thus, we modify the rotary position embedding as follows:
\begin{equation}
    \hat{\theta}_i = \begin{cases} 
    \frac{\theta_i}{s} & \text{if} \ \mathcal{D}^i(P^{\mathcal{E}}_{L'},P_{L}) > \mathcal{D}^i(P^{\mathcal{I}}_{L'},P_{L}) +t\\
    \theta_i & \text{otherwise},
    \end{cases}
    \label{equ:mddtheta}
\end{equation}
where $t$ is a threshold to determine the extension strategy when the disturbance scores $\mathcal{D}^i(P^{\mathcal{E}}_{L'},P_{L})$ and $\mathcal{D}^i(P^{\mathcal{I}}_{L'},P_{L})$ are very close. As demonstrated in \cref{equ:mddtheta}, for the $i$th dimension, we employ linear interpolation with $s_i=L'/L$, when its disturbance score is much smaller. Otherwise, direct extrapolation is a preferred choice for this dimension.

It's worth noting that our approach is a pre-execution strategy that does not add any time or calculation cost during the inference phase as long as the extension length $L'$ is provided.
Besides, since we only modify the value of $\theta$, any advanced method that influences the attention mechanism, such as FlashAttention \citep{flashattention,flashattention2}, is still compatible.



%% file: table/longbench_main.tex
\begin{table*}[!htbp]
    \centering
    \begin{tabular}{cccccccc}
    
    \hline
    \textbf{Base} & \textbf{Model} & \textbf{Context} & \multicolumn{3}{c}{\textbf{Evaluation Context Length}} & \multicolumn{2}{c}{\textbf{Average}} \\
    \textbf{LLM} & \textbf{Name} & \textbf{Window} & \textbf{\ \ 0-4k \ \ } & \textbf{ \ 4-8k \  } & \textbf{ \ \ 8k+\ \ } & \textbf{Avg.} & \textbf{Avg.$_{>4k}$}\\
    \hline
     \multirow{6}*{LLaMA2-7B} & Original & 4k & 27.69 & 26.24 & 25.79 & 26.57 & 26.02 \\
     &PI(s=2) & 8k &28.21&26.90&26.79& 27.30&26.85 \\
    &PI(s=4) & 16k &29.46&29.53&27.59& 28.87&28.56 \\
    &YaRN(s=2) & 8k &27.99&27.01&26.93& 27.31&26.97 \\
    &YaRN(s=4) & 16k &27.92&29.19&28.85& 28.65&29.02 \\
    &CLEX(ms=16) & 64k &25.22 & 28.87 & 28.62 & 27.57 & 28.75 \\
    &Ours(s=2) & 8k &28.24&27.78&27.43& 27.82&27.61 \\
    &Ours(s=4) & 16k &\textbf{29.98}&\textbf{30.30}&\textbf{30.09}& \textbf{30.12}& \textbf{30.20} \\

    \hline
    \multirow{6}*{LLaMA2-13B} & Original & 4k & 26.97 & 26.05 & 26.27 & 26.43 & 26.16 \\
    &PI(s=2) & 8k &31.43&30.95&29.74&30.71&30.35 \\
    &PI(s=4) & 16k &30.80&31.33&30.86& 30.99&31.10 \\
    &YaRN(s=2) & 8k &31.00&30.42&30.07& 30.50&30.25 \\
    &YaRN(s=4) & 16k &31.59&31.35&29.89& 30.94&30.62 \\
    &CLEX(ms=16) & 64k &29.84 & 30.22 & 30.22 & 30.09 & 30.22 \\
    &Ours(s=2) & 8k &\textbf{31.64}&31.40&30.43& 31.16&30.91 \\
    &Ours(s=4) & 16k &31.58&\textbf{32.29}&\textbf{31.15}& \textbf{31.67}&\textbf{31.72} \\
    \hline
    \end{tabular}
    \caption{Comparative performance analysis of various context window extension methods on the Longbench-E benchmark. Avg. denotes the average score across all lengths, while Avg.$_{>4k}$ represents the average score for lengths exceeding the pre-training length. The scaling factor of CLEX \citep{CLEX} is dynamic, "ms" denotes the maximum scaling factor, and we set the maximum scaling factor to 16 in accordance with the settings of \citet{CLEX}.}
    \label{tab:longbench_main}
\end{table*}

%% file: section/5_experiment.tex
\section{Experiments}
In this section, we evaluate our distribution-based method on both long- and short-context benchmarks. The results show that models employing our method outperform existing context window extension methods, indicating a better context window extension of RoPE-based LLMs while maintaining their original short-context capabilities.

\input{table/huggingface_main}

\subsection{Experimental Details}
We validate the effectiveness of our method on the trending LLaMA2 \citep{llama2} model, including 7B and 13B parameter models. All models are trained on a subset of PG19 \citep{pg19} datasets. For $s=2$, models are fine-tuned for 1000 steps with a global batch size of 64 and max length of 8192. For $s=4$, models are fine-tuned for 500 steps with a global batch size of 64 and a max length of 16384. We set the default value of $b$ in ~\cref{equ:bucket} to 360. By adjusting the value of $t$ in ~\cref{equ:mddtheta}, we set the default number of interpolated dimensions to 80 for 8k extension and to 64 for 16k extension. See more details in \cref{sec:appendixB.1}.
\subsection{Long Context Evaluation}
To evaluate the model's capabilities on real-world long context tasks with an extended context window. We utilize the Longbench-E benchmark \citep{longbench}, which is specifically designed for evaluating models with long context window. The Longbench-E benchmark consists of 13 diverse tasks, with the average length of most tasks ranging from 5k to 15k. Furthermore, \citet{longbench} categorizes the test samples into groups based on length intervals of 0-4k, 4-8k, and 8k+ to provide an analysis of the model's performance variations at different input lengths.

~\Cref{tab:longbench_main} shows a side-by-side comparison of the LLaMA2 model extended from 4k to the context length of 8k and 16k via PI \citep{PI}, YaRN \citep{yarn} and our method. We observe that models of different parameter sizes, employing our method as the extension method, achieve optimal average results when extended to various context lengths. Compared to PI, our method achieves an average score improvement of up to 4.33\% when extending the context window of LLaMA2-7B to 16k. To further demonstrate the model's performance when surpassing the pre-training length, we also report the average scores for evaluations with lengths greater than 4k. When extended to 16k, we can observe that models using our method maintain their performance in the extended context length range, whereas the model employing PI exhibits performance degradation at the 7B model and YaRN exhibits performance degradation at the 13B model. We also evaluated the perplexity of the models as well as their performance on the RULER benchmark \citep{RULER}, as shown in Appendix ~\ref{sec:appendixB.2}.

\subsection{Short Context Validation}
We further evaluate the LLaMA2 models on the standard short context benchmark from the Hugging Face Open LLM Leaderboard \citep{huggingface} to observe how its ability in the original length range changes after extending the context window. Specifically, we use 0-shot TruthfulQA \citep{TruthfulQA} and Hellaswag \citep{HellaSwag}, 5-shot MMLU \citep{mmlu} and 25-shot ARC-c \citep{arc}. The results demonstrate that the performance using our method to extend the context window is not significantly affected.

As illustrated in \Cref{tab:longbench_main}, when extending the LLaMA2-7B model to 8k with our approach, we observe only a 0.12 average score decrease compared to the original model. Meanwhile, extending the context window of the LLaMA2-7B model to 16k using YaRN results in a maximum average performance drop of 0.53, which is further exacerbated in the case of PI. When applying our method to extend the context window of the LLaMA2-13B model, we can even achieve a slightly average performance improvement, suggesting that extending the model's context window with our method does not substantially harm the model’s capability.

\begin{figure}[t]
    \centering
    \includegraphics[width=1\linewidth,height=!]{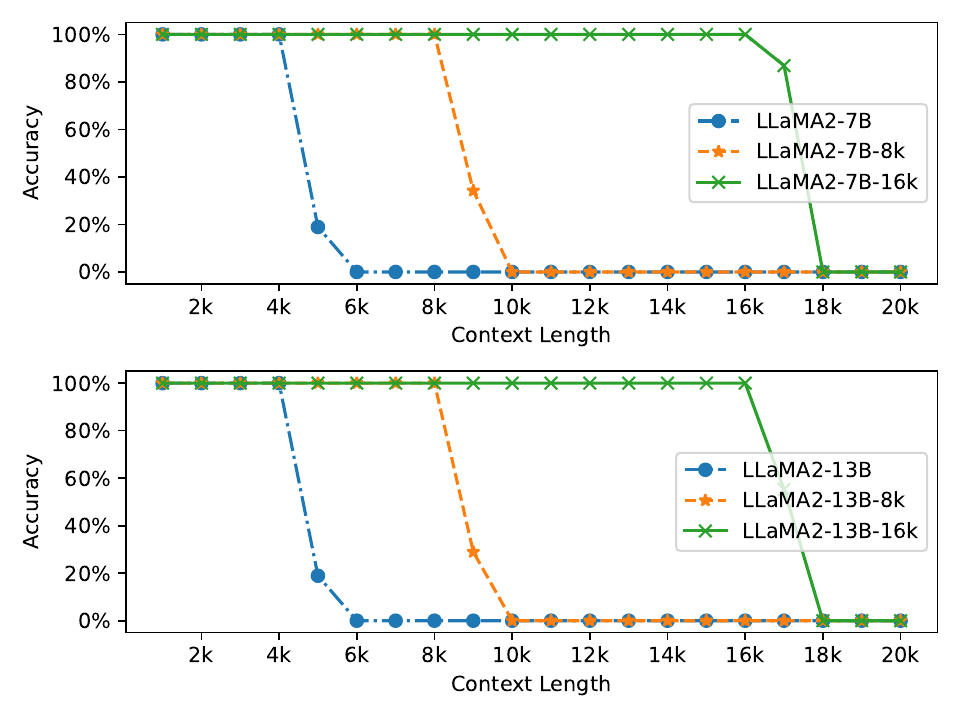}
    \caption{Passkey retrieval performance of models with different sizes under various context window lengths.}
    \label{passkey}
\end{figure}
\subsection{Passkey Retrieval}
To study the effective context window size of our model after extension, i.e. the maximum distance of a token that can be effectively attended to during inference. We further evaluate the model’s ability to retrieve a simple passkey from a massive amount of text via passkey retrieval task \citep{passkey}. 
Following the experimental setup of \citet{passkey}, we set the maximum input length for all models to 20k, with prompt details demonstrated in \Cref{sec:appendixB.3}. As shown in \Cref{passkey}, the LLaMA2 models, utilizing our context window extension approaches, achieve 100\% accuracy within the predetermined length.


%% file: table/huggingface_main.tex
\begin{table*}[t]
    \centering
    \begin{tabular}{cccccccc}
    \hline
    \textbf{Model} & \textbf{Model} & \textbf{Context} & \multirow{2}*{\textbf{TruthfulQA}} & \multirow{2}*{\textbf{Hellaswag}} & \multirow{2}*{\textbf{MMLU}} & \multirow{2}*{\textbf{ARC-c}} & \multirow{2}*{\textbf{Avg.}} \\
    \textbf{Name} & \textbf{Size} & \textbf{Window} \\
    \hline
    LLaMA2-7B & 7B & 4k &  38.74 & \textbf{77.38} & 46.96 & \textbf{52.22} & \textbf{53.82} \\
    \hline
    PI(s=2) & 7B & 8k & 38.03 & 76.61 & 44.02 & 50.68 & 50.35 \\
    PI(s=4) & 7B & 16k & 35.99 & 76.08 & 45.26 & 49.74 & 51.77 \\
    YaRN(s=2) & 7B & 8k & 39.10 & 76.83 & 46.05 & 51.45 & 53.36 \\
    YaRN(s=4) & 7B & 16k & 38.90 & 77.10 & 45.98 & 51.19 & 53.29 \\
    Ours(s=2) & 7B & 8k & \textbf{39.92} & 76.80 & 46.18 & 51.88 & 53.70 \\
    Ours(s=4) & 7B & 16k & 39.83 & 76.91 & \textbf{46.96} & 51.45 & 53.79 \\
    \hline
    LLaMA2-13B & 13B & 4k & 37.37 &\textbf{80.83} & 59.70 & 64.25 & 60.54 \\
    \hline
    PI(s=2) & 13B & 8k & 37.68 & 80.25 & 59.44 & 63.99 & 60.34 \\
    PI(s=4) & 13B & 16k & 35.35 & 79.94 & 58.76 & 61.77 & 58.96 \\
    YaRN(s=2) & 13B & 8k & 37.71 & 80.31 & 59.99 & 64.59 & 60.65 \\
    YaRN(s=4) & 13B & 16k & 38.53 & 80.35 & 59.05 & 64.16 & 60.52 \\
    Ours(s=2) & 13B & 8k & 38.10 & 80.09 & \textbf{60.16} & \textbf{64.68} & \textbf{60.76} \\
    Ours(s=4) & 13B & 16k & \textbf{39.26} & 80.03 & 59.57 & 64.08 & 60.74 \\
    \hline
    \end{tabular}
    \label{tab:huggingface_main}
    \caption{Comparative performance of various context window extension methods relative to the original LLaMA2 on the Hugging Face Open LLM benchmark.}
\end{table*}

%% file: section/6_analysis.tex
\section{Analysis}
In this section, we analyze the impact of distributional disturbance on model performance. Moreover, we analyze the selection of different interpolation dimension numbers in \cref{equ:mddtheta} and the impact of the number of intervals in \cref{equ:bucket}. All analyses are based on the task of extending the context window of LLaMA2-13B from 4k to 8k.

\subsection{Influence of Disturbance}
We calculate the distributional disturbance induced by different methods with \cref{equ:distance}. As illustrated in \Cref{tab:distance}, we achieve the lowest distributional disturbance, which is inline with experiment results.
\input{table/distance}
\begin{figure}[t]
    \centering
    \includegraphics[width=1\linewidth]{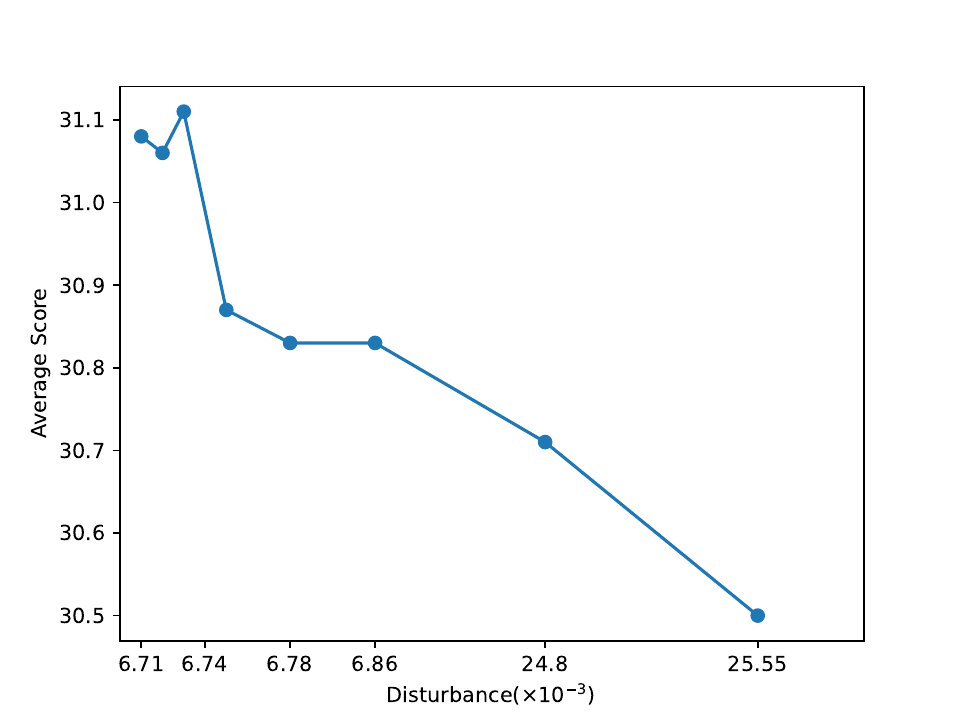}
    \caption{Performance of LLaMA2 declines on the LongBench-E with the increasing disturbance.}
    \label{fig:distance}
\end{figure}

Furthermore, when extending the context window of LLaMA2-13B to 8k, we investigate the model's extension performance with increased disturbance via incrementing the value of $t$ in \cref{equ:mddtheta}. As shown in \Cref{fig:distance}, with the disturbance increases, the performance of the model basically shows a monotonically decreasing trend, which reveals a strong consistency between the disturbance metric and the experimental performance.

\input{table/t_longbench}
\input{table/t_huggingface}
\subsection{Influence of Interpolation Dimension}
Let us denote the number of interpolation dimensions as $0\leq\hat{n}\leq d$.
In \cref{equ:mddtheta}, we can control the value of $t$ to decide how many dimensions the interpolation strategy is used for.
We demonstrate the influence of the number of interpolated dimensions $\hat{n}$ in \Cref{tab:t_longbench}, where $\hat{n}$ decreases from 96 to 56 as $t$ increases.
We observe that for dimensions where the disturbance scores $\mathcal{D}^i(P^{\mathcal{E}}_{L'},P_{L})$ and $\mathcal{D}^i(P^{\mathcal{I}}_{L'},P_{L})$ are very close, corresponding to the cases of $\hat{n}$ = 96, 88, and 80, 
the impact of choosing extrapolation or interpolation on the model's performance is slight and negligible. However, as the disturbance increases, corresponding to the cases of $\hat{n} < \text{80}$, maintaining distributional consistency becomes crucial, and we can observe a gradual decline in the performance when employing extrapolation to those dimensions where the disturbance score $\mathcal{D}^i(P^{\mathcal{E}}_{L'},P_{L})$ is significantly larger than $\mathcal{D}^i(P^{\mathcal{I}}_{L'},P_{L})$. We illustrate the influence of interpolation dimension numbers on downstream tasks in \Cref{tab:t_huggingface}, where the value of $\hat{n}$ has little effect and different datasets prefer different $\hat{n}$.

\subsection{Influence of Interval}
During the analysis of the rotary angle distribution in \cref{equ:bucket}, we divide $[0,2\pi)$ into $b$ intervals and statistically estimate their distribution. In this part, we explore the impact of $b$, ranging from $90$ to $720$, on the extension of the model's context window. As shown in \Cref{tab:bucket_longbench}, when $b$ = 90, 180 and 360, the model's performance after extension exhibits no significant fluctuations. This suggests that the model is capable of tolerating subtle differences in rotation angles.
The performance drops when $b=720$. This is because excessive intervals can actually increase the error in the distribution estimation, since the number of rotary angle samples $L$ is not very large. \Cref{tab:bucket_huggingface} illustrates that the choice of $b$ does not influence the downstream tasks.
\input{table/bucket_longbench}
\input{table/bucket_huggingface}


%% file: table/distance.tex
\begin{table}[t]
    \centering
    \begin{tabular}{c|cc}
    \hline
    \multirow{2}*{\textbf{Method}} & \multicolumn{2}{c}{\textbf{Context Length}} \\
    & 8k & 16k \\
    \hline
    PI & 24.08 & 33.67 \\
    YaRN & 25.55 & 35.44 \\
    Ours & \textbf{6.71} & \textbf{22.92}\\
    \hline
    \end{tabular}
    \caption{Disturbance($\times 10^{-3}$) of rotary angle distributions resulting from difference methods when extended to various length. Our method has the lowest disturbance. More details are shown in \cref{sec:appendixA.2}.}
    \label{tab:distance}
\end{table}

%% file: table/t_longbench.tex
\begin{table}[t]
    \centering
    \begin{tabular}{ccccc}
    \hline
    \textbf{$\hat{n}$} & \textbf{0-4k} & \textbf{4-8k} & \textbf{8k+} & \textbf{Avg.} \\
    \hline
    56 & 31.20 & 31.07 & 29.87 & 30.71 \\
    64 & 31.19 & 31.03 & 30.27 & 30.83 \\
    72 & 31.39 & 31.12 & 30.12 & 30.87 \\
    80 & \textbf{31.64} & 31.40 & 30.43 & \textbf{31.15} \\
    88 & 31.32 & 31.34 & \textbf{30.53} & 31.06 \\
    96 & 31.38 & \textbf{31.52} & 30.34 & 31.08 \\
    \hline
    \end{tabular}
    \caption{Influence of interpolation dimension numbers $\hat{n}$ on the long context benchmark.}
    \label{tab:t_longbench}
\end{table}

%% file: table/t_huggingface.tex
\begin{table}[t]
    \setlength\tabcolsep{2pt}
    \centering
    \begin{tabular}{ccccc}
    \hline
    \textbf{$\hat{n}$} & \textbf{TruthfulQA} & \textbf{Hellaswag} & \textbf{MMLU} & \textbf{ARC-c} \\
    \hline
    56 & \textbf{38.95} & \textbf{80.27} & 60.29 & 64.25 \\
    64 & 38.68 & 80.23 & 60.55 & \textbf{64.76} \\
    72 & 38.51 & 80.27 & 60.22 & 64.16 \\
    80 & 38.10 & 80.09 & 60.16 & 64.68 \\
    88 & 37.74 & 80.17 & \textbf{60.61} & \textbf{64.76} \\
    96 & 38.60 & 80.14 & 60.09 & \textbf{64.76} \\
    \hline
    \end{tabular}
    \caption{Influence of interpolation dimension numbers $\hat{n}$ on the Hugging Face Open LLM benchmark.}
    \label{tab:t_huggingface}
\end{table}

%% file: table/bucket_longbench.tex
\begin{table}[t]
    \centering
    \begin{tabular}{ccccc}
    \hline
    \textbf{$b$} & \textbf{0-4k} & \textbf{4-8k} & \textbf{8k+} & \textbf{Avg.} \\
    \hline
    90 & 31.47 & 31.26 & 30.44 & 31.06 \\
    180 & \textbf{31.68} & 31.09 & \textbf{30.51} & 31.09 \\
    360 & 31.64 & \textbf{31.40} & 30.43 & \textbf{31.15} \\
    720 & 31.32 & 31.03 & 30.18 & 30.84 \\
    \hline
    \end{tabular}
    \caption{Influence of the interval numbers $b$ on the long context benchmark.}
    \label{tab:bucket_longbench}
\end{table}

%% file: table/bucket_huggingface.tex
\begin{table}[t]
    \setlength\tabcolsep{2pt}
    \centering
    \begin{tabular}{ccccc}
    \hline
    \textbf{$b$} & \textbf{TruthfulQA} & \textbf{Hellaswag} & \textbf{MMLU} & \textbf{ARC-c} \\
    \hline
    90 & 37.44 & 80.12 & \textbf{60.74} & 64.68 \\
    180 & 38.18 & 80.23 & 60.48 & \textbf{64.76} \\
    360 & 38.10 & 80.09 & 60.16 & 64.68 \\
    720 & \textbf{38.78} & \textbf{80.29} & 60.35 & 64.33 \\
    \hline
    \end{tabular}
    \caption{Influence of the interval numbers $b$ on the Hugging Face Open LLM benchmark.}
    \label{tab:bucket_huggingface}
\end{table}

%% file: section/8_relatedworks.tex
\section{Related Works}
Long-sequence modeling is a crucial issue in the application of LLMs. Recent efforts focus on improving position embedding to enable LLMs have larger context window. Currently, the most popular relative position embedding are ALiBi \citep{alibi} and RoPE \citep{rope}. ALiBi \citep{alibi} adds bias to attention, enabling models to maintain lower perplexity on long sequences, but only generalizes to limited lengths on downstream tasks \citep{alibit}. RoPE \citep{rope} cannot generalize to lengths beyond its pre-training length.

Some works have been done to overcome such limitation. \citet{randomrope} randomize token's position embedding during pre-training, enabling the model based on RoPE to generalize to predetermined sequence lengths. This effectively guarantees consistency in the distribution of rotation angles when generalizing to predetermined lengths, demonstrating that rotation angle distribution consistency is crucial for the model's ability to generalize. \citet{PI,fixedNTK,dynamicNTK, scalinglaw, yarn} extend the context window of existing LLMs (i.e., LLaMA2 \citep{llama2}) by slightly modifying RoPE's $\theta$ (as show in ~\cref{equ:theta}). \citet{PI} achieves proposed to extend the context window by interpolating positions, using a scaling factor $s=L'/L$ to uniformly scale $\theta_i$, and fine-tuning on a small amount of data to extend the model's context window. \citet{fixedNTK, dynamicNTK} base on the Neural Tangent Kernel (NTK) theory, they scale lower dimensions less and higher dimensions more, this is also referred to as Adjusted Base Frequency (ABF). \citet{scalinglaw} achieves an effect similar to NTK by modifying the base of RoPE. YaRN \citep{yarn} improved NTK by dividing RoPE dimensions into three frequency-based groups and applying different strategies to each group. Low frequency ($r_i<\alpha$) dimensions use interpolation like PI and high frequency ($r_i>\beta$) dimensions use extrapolation, dimensions that fall in-between employs the NTK. YaRN achieved good performance, but lacked interpretability, the hyperparameters $\alpha$ and $\beta$ were also empirically chosen, making it hard to obtain the optimal results. Different from these empirical method, our work initially highlights the consistency of rotary angle distribution as a theoretical guidance for extending the context window. 

%% file: section/7_conclusion.tex
\section{Conclusion}
In this work, we proposed to study the context window extension from a distributional perspective and demonstrated that the consistency of rotary angle distributions has a significant impact on extending the context window of LLMs based on the rotary position embedding.
We designed a framework to select scaling strategies with the guidance of minimizing the disturbance of rotary angle distributions.
Experimental results demonstrated the effectiveness and superiority of our approach.
Although our approach is limited by the rotary position embedding, we believe that our distributional perspective has the potential to inspire future work.

%% file: table/ruler.tex
\begin{table*}[ht]
    \centering
    \begin{tabular}{ccccccc}
    
    \hline
    \textbf{Base} & \textbf{Model} & \textbf{Context} & \multicolumn{3}{c}{\textbf{Evaluation Context Length}} & \multirow{2}*{\textbf{Avg.}} \\
    \textbf{LLM} & \textbf{Name} & \textbf{Window} & \textbf{\ \ 4k \ \ } & \textbf{ \ 8k \  } & \textbf{ \ \ 16k\ \ } & \\
    \hline
     \multirow{6}*{LLaMA2-7B} & Original & 4k &\textbf{82.23}&0&0& 27.41 \\
    &PI(s=4) & 16k &75.22&72.61&68.81& 72.21 \\
    &YaRN(s=4) & 16k &76.21&72.84&67.70& 72.25\\
    &CLEX(ms=16) & 64k & 53.04 & 49.38 & 49.79 & 50.74 \\
    &Ours(s=4) & 16k & 78.74 & \textbf{75.55} & \textbf{71.78} & \textbf{75.35}\\
    
    \hline
    \multirow{6}*{LLaMA2-13B} & Original & 4k & \textbf{84.93}& 0 & 0 & 28.31\\
    &PI(s=4) & 16k &76.22&72.41&66.97& 71.87 \\
    &YaRN(s=4) & 16k &72.37&68.97&63.27& 68.20 \\
    &CLEX(ms=16) & 64k & 58.27 & 53.69 & 51.48 & 54.48 \\
    &Ours(s=4) & 16k &79.40& \textbf{76.21} & \textbf{71.65}& \textbf{75.75}\\
    \hline
    \end{tabular}
    \caption{Comparative performance analysis of various context window extension methods on the RULER benchmark. The scaling factor of CLEX is dynamic, "ms" denotes the maximum scaling factor, and we set the maximum scaling factor to 16 in accordance with the settings of \cite{CLEX}.}
    \label{tab:ruler_main}
\end{table*}

%% file: table/time.tex
\begin{table}[h]
    \centering
    \begin{tabular}{cccc}
    \hline
    \textbf{Model Size} & \textbf{Method} & \textbf{Train} & \textbf{Test} \\
    \hline
    \multirow{4}*{7B} 
    &PI & 1 & 1 \\
    &YaRN & 1 & 1 \\
    &CLEX & 1.62 & 1.83 \\
    &Ours & 1 & 1\\
    \hline
    \multirow{4}*{13B}
    &PI & 1 & 1 \\
    &YaRN & 1 & 1 \\
    &CLEX & 1.53 & 1.81 \\
    &Ours & 1 & 1\\
    \hline
    \end{tabular}
    \caption{Time cost of diferent methods.}
    \label{tab:time}
\end{table}

%% file: table/ppl.tex
\begin{table}[h]
    \centering
    \begin{tabular}{cccc}
    \hline
    \multirow{2}*{\textbf{Model Size}} & \multirow{2}*{\textbf{Method}} & \multicolumn{2}{c}{\textbf{Context Length}} \\
    & & 8k & 16k \\
    \hline
    \multirow{4}*{7B} 
    &PI & 8.19 & 9.35 \\
    &YaRN & 7.39 & 7.82 \\
    &CLEX & 7.30 & 7.87 \\
    &Ours & \textbf{7.12} & \textbf{7.72}\\
    \hline
    \multirow{4}*{7B}
    &PI & 7.02 & 8.23 \\
    &YaRN & 6.06 & 7.77 \\
    &CLEX & 6.08 & 7.58 \\
    &Ours & \textbf{5.91} & \textbf{7.39}\\
    \hline
    \end{tabular}
    \caption{Sliding window perplexity (S = 256) on PG19 dataset.}
    \label{tab:ppl}
\end{table}